\documentclass{article}
\usepackage{spconf,amsmath,epsfig}

\usepackage{amsfonts,amssymb}
\usepackage{overpic}
\usepackage{graphicx}
\usepackage{subcaption,siunitx,booktabs}
\usepackage[colorlinks=true, allcolors=blue]{hyperref}
\usepackage[ruled,vlined,linesnumbered,noresetcount]{algorithm2e}
\usepackage{amsmath}
\usepackage{multicol}
\usepackage{multirow}
\usepackage{mathtools}
\usepackage[title]{appendix}
\usepackage{float}

\usepackage[inkscapelatex=false]{svg}

\DeclareMathOperator{\NN}{NN}
\DeclareMathOperator{\ODESolve}{ODESolve}

\SetKwInOut{Parameter}{Parameters}
\SetKwFunction{ODESolveAlg}{ODESolve}
\SetKwFunction{Forward}{FwdPass}
\SetKwFunction{concat}{Concat}
\SetKwFunction{criterion}{Criterion}

\title{Graph Neural Differential Equations for Coarse-Grained Socioeconomic Dynamics}

\name{
\begin{tabular}{@{}c@{}}
James Koch$^{\star}$ \qquad 
Pranab K. Roy Chowdhury$^{\star}$ \qquad 
Heng Wan$^{\star}$ \\
\end{tabular}
\\
\begin{tabular}{@{}c@{}}
Parin Bhaduri$^{\dagger}$ \qquad 
Jim Yoon$^{\star}$ \qquad 
Vivek Srikrishnan$^{\dagger}$ \qquad
W. Brent Daniel$^{\star}$  \\
\end{tabular}
\thanks{This work was supported by the Multisector Dynamics program area of the U.S. Department of Energy, Office of Science, Office of Biological and Environmental Research as part of the multi-program, collaborative Integrated Coastal Modeling (ICoM) project at PNNL, a multi-program national laboratory operated by Battelle for the U.S. Department of Energy under contract DE-AC05-76RL01830. \\
Contact: \href{mailto:james.koch@pnnl.gov}{james.koch@pnnl.gov}. }}

\address{$^{\star}$ Pacific Northwest National Laboratory, Richland, WA, USA \\
     $^{\dagger}$Cornell University, Ithaca, NY, USA}

\begin{document}

\maketitle

\begin{abstract}
We present a data-driven machine-learning approach for modeling space-time socioeconomic dynamics. Through coarse-graining fine-scale observations, our modeling framework simplifies these complex systems to a set of tractable mechanistic relationships -- in the form of ordinary differential equations -- while preserving critical system behaviors. This approach allows for expedited 'what if' studies and sensitivity analyses, essential for informed policy-making. Our findings, from a case study of Baltimore, MD, indicate that this machine learning-augmented coarse-grained model serves as a powerful instrument for deciphering the complex interactions between social factors, geography, and exogenous stressors, offering a valuable asset for system forecasting and resilience planning.
\end{abstract}

\section{Introduction}

\subsection{Modeling Socioeconomic Systems: ABMs}
Socioeconomic systems exhibit intricate patterns of interaction and adaptation, e.g., between social groups and to changing environmental and economic conditions, reflecting a degree of complexity that challenges current methods for analysis and prediction of these systems \cite{holling2001understanding, manson2001simplifying}. Modeling the complexity of these systems often requires detailed knowledge of the interacting components, their associated scales, and sufficient resolution of these details such that bottom-up emergent properties can be observed. Agent-based models (ABMs) are a computational modeling tool to resolve these facets of complex social science problems. ABMs simulate a large number of agents in a shared environment: through agent-to-agent and agent-environment interactions, ABMs relate system-wide emergent properties to individual behaviors \cite{macy2002factors}. 

ABMs represent one end of a spectrum of modeling strategies, where a practitioner can explicitly specify a model's expressivity through its resolution and logic. These choices are often manifest as the ABM's degrees of freedom and the rules for their evolution. The number of degrees of freedom is typically large, with each agent possessing time-varying attributes and mechanisms that dynamically respond to each agent's perception of its environment \cite{sun2016simple}. This allows ABMs to simulate complex systems with a high degree of detail, capturing emergent phenomena that arise from the collective behavior of individual agents. 

Despite their success across various subject domains and use cases, including computational social science, ABMs are not without limitations. They can suffer from prohibitive computational demands, challenges in defining appropriate agent rules and interaction environments, and difficulties in model validation and verification, particularly when used for predictive tasks \cite{crooks2008key, larsen2016appropriate}. 

\subsection{Model Coarse-Graining}
At the other end of the modeling spectrum are mean-field and coarse-grained models. These attempt to aggregate a system's degrees of freedom to extract, track, and evolve salient features. Mean-field models simplify many-body problems with effective `fields', significantly reducing model complexity \cite{ye2021bridging}. Coarse-grained models perform a similar simplification by aggregating to a user-chosen granularity; e.g. lumping a group of agents from an ABM into a single representative unit with far fewer degrees of freedom \cite{jin2022bottom}. These methods decrease model complexity while retaining interpretability that is relevant to the modeling objective, such as tracking population-wide metrics over time. The ability to choose a level of fidelity in a coarse-grained model is particularly useful in multi-scale modeling, where interactions across physical scales is important for system evolution, but difficult to fully resolve.

Coarse-graining comes with its own set of difficulties: while more tractable, pairing a model specification with a chosen granularity is not straightforward. For example, in an ABM, one parameterizes an agent's behaviors by a set of rules that best represent one's prior beliefs about an agent. If coarse-grained (e.g. averaging over a set of these agents), those same rules are not valid for the coarse-grained unit. New rules must be crafted to match the new granularity of the model. This is difficult to do in a from-first-principles fashion; often coarse-grained models take the form of an Equation-Based Model (EBM) where a system's aggregated states evolve according do a differential equation. 

\subsection{Machine Learning for Complex Systems}

Machine Learning (ML) has emerged as a powerful tool for modeling and scientific discovery of complex systems. With the proliferation of data (e.g. observational and simulation), an opportunity exists to apply these techniques in an effort to extract insights. These can range from data-driven analytics to calibrating mechanistic models given system observations. In the context of complex systems, such as our motivating example of socioeconomic dynamics of coastal urban development, we postulate that ML can be leveraged to identify governing dynamics, including feedbacks and nonlinearities. 

\subsubsection{ML across Social Systems}

The application of ML to social systems has accelerated in recent years. Applications range from finance to healthcare to real estate, to name a few domains. For instance, Culkin et al. \cite{culkin2017machine} implemented deep learning to learn to price options with excellent quantitative performance. Jiang et al. \cite{jiang2022application} implemented Support Vector Machine (SVM) and Artificial Neural Networks (ANN) on stock price forecasts, while Barboza et al. \cite{barboza2017machine} achieved a better prediction of bankruptcy than the traditional methods by using ML. For the healthcare domain, Doupe et al. \cite{doupe2019machine} investigated various ML techniques’ potential to predict healthcare outcomes, including cost, utilization, and service quality. In the real estate sector, ML has become essential for enhancing the accuracy of housing price predictions, as shown by Selim \cite{selim2009determinants}.  These examples demonstrate the extensive applications of ML across various domains, highlighting its capacity to enhance understanding and improve model performance in social systems.  

\subsubsection{Surrogate models for ABM}
Using ML methods to help alleviate the bottlenecks of ABMs has also been explored \cite{pietzsch2020metamodels}. Surrogate or meta-models have been proposed in several contexts as a vector for ABM emulation, sensitivity analysis, and the like. Learning the input-output space of an ABM using ANNs has been shown to be a promising method for surrogate modeling \cite{angione2022using}. Incorporating temporal dependencies in a mechanistic model has enabled near-real-time output \cite{kieu2024towards} for simulations of dynamics of human crowds. Using surrogate models to aid in calibration and sensitivity analysis has also been explored, such as in Lampert et al. \cite{lamperti2018agent}.

\subsubsection{Equation Discovery and Closure Models}
Machine learning can also be used to extract parsimonious relationships from data. In Zhang et al,. \cite{zhang2020data}, data from an ABM for epidemiology is aggregated into susceptible (S), infected (I), and recovered (R) groups that evolve as an EBM. In this form, conversion rates between the different sub-populations becomes an explicit model output and system forecasting can be performed with a greater degree of certainty. This work leverages the Sparse Identification of Nonlinear Dynamics (SINDy \cite{brunton2016discovering}) for the equation discovery task. In SINDy, a differential equation is constructed from a sparse selection of mathematical terms from a library of candidate expressions that minimizes a time series reconstruction error.  While powerful, such methods are limited by: (i) the noisiness of estimated temporal derivative, and (ii) the library, which needs to contain the `correct' terms needed to reconstruct the differential equation. The latter is encountered when observations are not of a system's state variable, not observed in intrinsic coordinates, or are ill-resolved. This is especially relevant in the context of social systems, where state variables, observables, features, etc., are ill-defined.

Neural Ordinary Differential Equations (Neural ODEs) are a similar ML technique that seek to approximate a time series with differential equations. Where SINDy uses a library of candidate functions, Neural ODEs leverage ANNs to approximate dynamics (e.g. $dx/dt = \NN(x,t)$). Interestingly, Neural ODEs can be extended to include prior domain knowledge by construction. The Universal Differential Equation modeling paradigm \cite{rackauckas2020universal} is that which includes known physical priors in the model equations. For example, one could model the speed $v$ of an object in freefall as $dv/dt = -9.81 \text{m/s\textsuperscript{2}} + \NN(v)$; that is, one knows the acceleration due to gravity, but not how the object interacts with air.

\subsection{Specific Aims and Contributions}

\begin{figure*}[]
        \centering
            \begin{overpic}[width=1.0\linewidth]{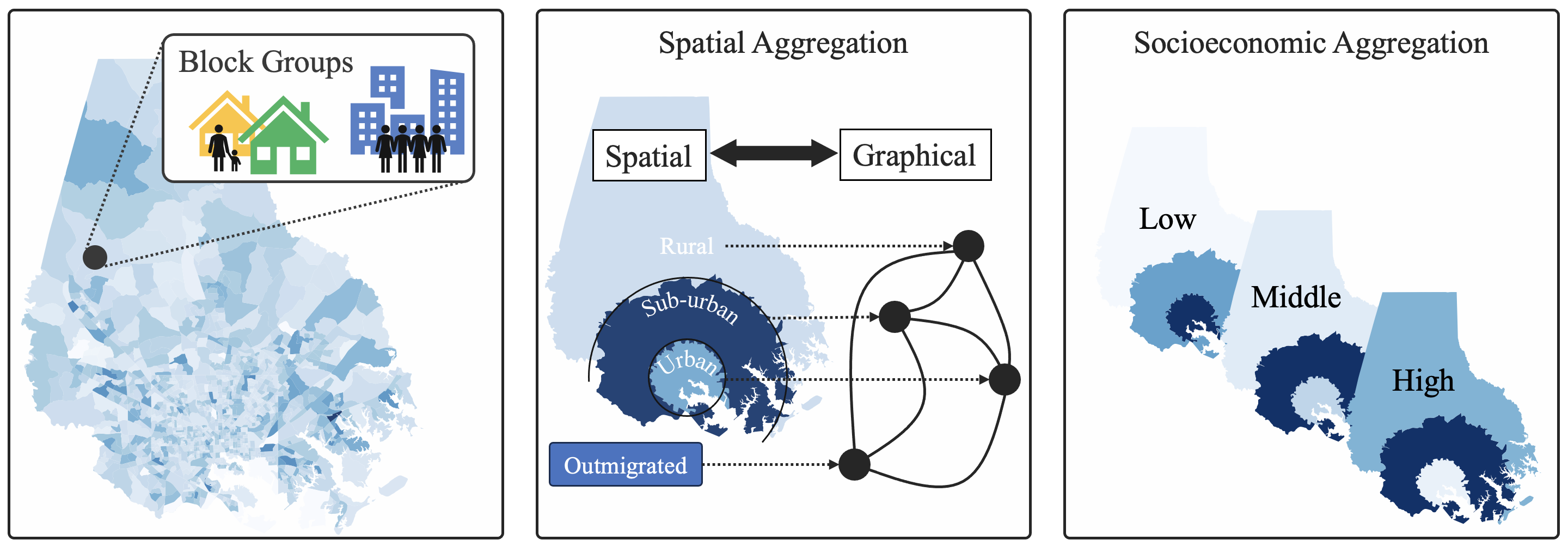}  
            \put(1,32){(a)}
            \put(34.75,32){(b)}
            \put(68.5,32){(c)}
	    \end{overpic}  
	    \caption{In (a), Balitimore County, MD is displayed as the composition of individual census block groups that contain people and housing. These data can be spatially aggregated (e.g. based on distance to the city center, as is depicted in (b)) into zones and equivalently represented as nodes on a graph. In this example, the edges between the nodes represent migration pathways. The population contained in these nodes can be similarly aggregated into socioeconomic groups; e.g. low, middle, and high income sub-populations, as is shown in (c).}
		\label{fig:fig_01}
\end{figure*}

We seek to model space-time socioeconomic dynamics from a data-driven perspective. Our goal is to create a mechanistic, coarse-grained equation-based model of full-fidelity mechanics, enabling faster simulations that retain model interpretability. 

Our specific contributions are:
\begin{itemize}
    \item The conceptualization and implementation of a scale-agnostic EBM for space-time socioeconomic systems, 
    \item Pairing an arbitrary coarse-graining procedure with a method to extract an EBM of the system `physics' at that model granularity,
    \item Performing this task in a differentiable manner, enabling gradient-based optimization of the coarse-grained EBM.
\end{itemize}

This paper is organized as follows: Section \ref{sec:methods} details our methods, complete with a discussion of graphs, Neural ODEs, and closure modeling. Our motivating case study is presented in Section \ref{sec:results}, followed by a discussion of the modeling approach in Section \ref{sec:discussion} and conclusions in Section \ref{sec:conclusion}

\section{Methods} \label{sec:methods}
In this section, we introduce and formalize the description of our proposed ML-enabled coarse-grained EBM. The EBM is inspired by widely-used bathtub-type models, where a quantity of interest (e.g. a bathtub's water level) is related to the inflow, outflow, and capacitance of a unit (e.g. a bathtub's cross-sectional area). These quantities can be related as an input-output accounting scheme in a differential equation: $\frac{dh}{dt} = \frac{1}{A}\left(Q_{\text{in}} - Q_{\text{out}}\right)$, where $h$ is the height of the water in the bathtub, $A$ is its area, and $Q$ denotes flow rates. In this simple model, the height of the water evolves according to the difference in inlet and outlet flows.

Here, we consider a network of such 'bathtubs', or nodes, that store people as opposed to a fluid. These nodes represent capacity to house subsets of a population, e.g. low, middle, and high-income groups. This is analogous to a bathtub filled by a mixture of different liquids. Continuing this analogy, a network is defined by many bathtubs connected by different sized pipes, or edges, that allow for the flow of fluid between them. 

Our model seeks to construct a network of capacitive nodes that represent geographic regions of an urban setting; e.g. a cluster of census block groups aggregated into a single effective unit (Fig \ref{fig:fig_01}). These nodes are connected by migration pathways along which populations flow. The model is a coarse-grained, EBM that tracks the flux of sub-populations between nodes over time. While the mathematics of networked capacitive nodes is well understood in our bathtub analogy (diffusion on a graph, Resistive-Capacitive networks, etc.), this is not true for our motivating problem. Urban socioeconomic dynamics are not well understood nor characterized from a first-principles perspective. 

\subsection{Definitions and Preliminaries}
\subsubsection{Graphs}
A graph is comprised of nodes and the edges that connect them. We denote $\mathcal{V}$ as the set of nodes in a graph and $\mathcal{E}$ as the set of edges. The number of nodes in the graph is $|\mathcal{V}| = n$. An edge $\epsilon \in \mathcal{E}$ is defined by a pair of elements of $\mathcal{V}$; i.e.  $\epsilon_{i,j} = \{\nu_i, \nu_j\}$ and $\nu_i, \nu_j \in \mathcal{V}$. A graph $\mathcal{G}$ is therefore defined as $\mathcal{G} = \{\mathcal{V}, \mathcal{E}\}$. The adjacency matrix $A$ of the graph $\mathcal{G}$ is a representation of all edges and nodes in a graph; if an edge $\epsilon_{i,j}$ connects two nodes, that entry of $A$ is 1:
\begin{equation}
A_{i,j}=\begin{cases}
			1, & \text{if $\epsilon_{i,j} \in \mathcal{E}$}\\
            0, & \text{otherwise}
		 \end{cases} .
\end{equation}
In our work, nodes contain time-varying states and features. States are  variables to be evolved according to the EBM (e.g. the composition of subpopulations at a node). Features are derived quantities or other attributes that further describe the nodes, but do not evolve according to the governing equation (e.g. distance to central business district, housing capacity, etc.). We denote the $n_x$-dimensional state vector for each node $\nu$ as $\mathbf{x}_{\nu} \in \mathbb{R}^{n_x}$ and the corresponding $n_y$-dimensional feature vector as $\mathbf{y}_{\nu} \in \mathbb{R}^{n_y}$.

\subsubsection{Neural Networks}
Artificial Neural Networks (ANNs) are trainable universal function approximators. An ANN consists of layers of interconnected nodes. At a minimum, an ANN is a composition of an input layer, one or more hidden layers, and an output layer.
The input layer receives the initial data, which is then processed through the hidden layers using a series of weighted connections. Each neuron in the hidden layers computes a weighted sum of its inputs and applies a non-linear activation function. The process can be represented mathematically for a single neuron as:
\begin{equation}
    a_j = \phi\left(\sum_{i=1}^{n} w_{ij} x_i + b_j \right)
\end{equation}
where $x_i$ are the input features, $w_{ij}$ are the layer weights, $b_j$ is the bias term, $\phi$ is the activation function (e.g. hyperbolic tangent), and $a_j$ is the neuron's output. In this work, we use the notation $\NN(x) : \mathbb{R}^{n_x} \rightarrow \mathbb{R}^{\text{out}}$ to represent a complete neural network with a prescribed input-output mapping. Given a modeling objective and loss function, the network's weights are optimized via the backpropagation algorithm (gradient-based optimization). 

\subsubsection{Neural Ordinary Differential Equations}
Neural Ordinary Differential Equations (Neural ODEs \cite{chen2018neural}) provide a framework for modeling continuous-time dynamics of a system using neural networks. Consider a dynamical system characterized by a time-dependent state vector $\mathbf{x}(t) \in \mathbb{R}^d$: 
\begin{equation} \frac{d \mathbf{x}}{dt} = f(\mathbf{x}, t; \theta), \end{equation}
where $f: \mathbb{R}^d \times \mathbb{R} \times \Theta \rightarrow \mathbb{R}^d$ is a neural network parameterized by $\theta \in \Theta$.  The ANN maps the current state and time to the rate of change of the state. The trajectory of the system from an initial time $t_0$ to a final time $t_{end}$ is obtained by solving the initial value problem:
\begin{equation} \mathbf{x}^{(t_{end})} = \mathbf{x}^{(t_0)} + \int_{t_0}^{t_{end}} f(\mathbf{x}, t; \theta) dt, \end{equation}
which, in practice, is computed using a numerical ODE solver:
\begin{equation} \mathbf{x}^{(t_{end})} = \text{ODESolve}(f, \mathbf{x}^{(t_0)}, t_0, t_{end}, \theta). \end{equation}
To fit the Neural ODE to observed data, the parameters $\theta$ of the function $f$ are adjusted to minimize a loss function that quantifies the discrepancy between the model's predictions and the true data. This optimization is performed using gradient-based methods, leveraging the differentiability of the ODE solver with respect to $\theta$. The gradients can be computed either by backpropagating through the operations of the numerical solver (reverse-mode automatic differentiation) or by solving an adjoint sensitivity problem that efficiently computes gradients with respect to the initial conditions and parameters. These techniques are implemented in modern software libraries such as DiffEqFlux.jl \cite{rackauckas2019diffeqflux} for Julia and torchdiffeq for Python, which facilitate the integration of Neural ODEs into machine learning pipelines.

\subsection{Population Flux on a Graph}
We are interested in modeling aggregated population dynamics on a graph. The nodes of a graph correspond to spatially aggregated portions of a domain, each equipped with a notion of population capacity, occupancy rate, and a list of attributes (e.g. proximity to central business district, flood inundation depth, etc.), as depicted in Fig. \ref{fig:fig_02}. In our bathtub analogy, fluid moves through the network according to pressure gradients: so long as a difference in fluid height exists, the fluid will continue to seek height equilibrium through the network, subject to sources, sinks, and the network topology.  In this work, we similarly define a notion of `pressure' experienced by these sub-populations on the nodes and how each unique sub-population responds to pressure differences on the graph. Intuitively, one expects populations to migrate towards regions of lower pressure. However, different socioeconomic groups may perceive identical conditions in a node as having vastly different `pressures,' as is noted in instances of segregation \cite{schelling1971dynamic}. Similarly, not all populations flow according to a negative pressure gradient. This section introduces these concepts as `placeholders' to be explicitly defined in our motivating use case in Section \ref{sec:results}.

Let $n = |\mathcal{V}|$ be the number of a model's spatially aggregated nodes and $n_x$ be the number of sub-populations. Each node $\nu$ contains a state variable for each sub-population; $\mathbf{x}_\nu \in \mathbb{R}^{n_x}$. Each node also contains a $n_y$-dimensional feature vector $\mathbf{y}_\nu \in \mathbb{R}^{n_y}$. We denote the full system state as the set:
\begin{equation}
    X = \{\mathbf{x}_{\nu_1}, \mathbf{x}_{\nu_2}, \dots ,\mathbf{x}_{\nu_n} \},
\end{equation}
which contains a total of $n_z \times n_x$ states. Similarly, we define the full system feature set as: 
\begin{equation}
    Y = \{\mathbf{y}_{\nu_1}, \mathbf{y}_{\nu_2}, \dots ,\mathbf{y}_{\nu_n} \}.
\end{equation}

On each node, the time rate of change of each sub-population is equal to the net flux into the node. Contributions to flux include networked migration and exogenous growth and decay (e.g. external in-migration). This can be written as:
\begin{equation} \label{eq:diffusion}
\begin{split}
    \underbrace{\frac{d \mathbf{x}_{\nu_i}}{dt}}_{\substack{\text{Time rate} \\ \text{of change}}} = &\underbrace{\sum_{\substack{j=1}}^{n} A_{i,j} \phi\left( \mathbf{y}_{\nu_j} - \mathbf{y}_{\nu_i} ;\theta \right) \beta\left(\mathbf{x}_{\nu_i},\mathbf{x}_{\nu_j}\right)}_{\text{Scaled node-to-node flux}} \\
    & + \underbrace{G_{\nu_i}\left(t\right)}_{\text{Exogenous growth}} - \underbrace{D_{\nu_i}\left(t\right)}_{\text{Exogenous decay}} \\
    & \eqcolon f_{\nu_i}\left(X, Y, t; \theta \right),
\end{split}
\end{equation}
where $\phi :  \mathbb{R}^{n_y} \times \Theta \rightarrow  \mathbb{R}^{n_x}$ maps differences in nodal features to nodal `pressure' with parametric dependence on $\theta$, $\beta: X \times X \rightarrow  \mathbb{R}^{n_x}$ is a scaling function, and $G_{\nu_i}:  \mathbb{R}^{1} \rightarrow \mathbb{R}^{n_x}$ and $D_{\nu_i}: \mathbb{R}^{1} \rightarrow \mathbb{R}^{n_x}$ are node-specific exogenous growth and decay terms, respectively. For notational convenience, we lump the right-hand-side of Eqn. \ref{eq:diffusion} as $f_{\nu_i} : X \times Y \times \mathbb{R} \times \Theta \rightarrow \mathbb{R}^{n_x}$. The full-system model is therefore:
\begin{equation} \label{eq:system}
    \frac{dX}{dt}  = \begin{pmatrix}
    f_{\nu_1}\left(X, Y, t; \theta\right)\\
    \vdots\\
    f_{\nu_n}\left(X, Y, t; \theta\right)
  \end{pmatrix} \eqcolon F\left(X, Y, t; \theta\right),
\end{equation}
where we have further simplified the notation for the system dynamics to $F: X \times Y \times \mathbb{R} \times \Theta \rightarrow \mathbb{R}^{n_z \times n_x}$. 

\subsection{Closure Problem}
Our goal is to fit Eqn. \ref{eq:diffusion} to time series observations of population dynamics on a graph. The multi-scale nature of the problem precludes our ability to match the granularity of the model with a from-first-principles `pressure' response function $\phi$. The closure problem is approximating these missing physics given some data. This can be done by imposing a particular functional form for $\phi$ (e.g. if a modeling practitioner had prior knowledge of what this term should look like) or by approximating it via an ANN.

Here, we assume we have selected a functional form for $\phi$, which could be a neural network, that has a parametric dependence on $\theta$ (we describe our modeling choices for our motivating case study in Section \ref{sec:results}). To optimize $\theta$, we define a closure problem. Let $M$ be a dataset containing $i \in I$ coarse-grained ABM simulation outputs. Each $M_i \in M$ is a time series of these coarse-grained states:
$$
   M_i \coloneq \{X^{\left(0\right)}_i, \dots, X^{\left(N\right)}_i \}.
$$
Here, the superscript in parentheses is understood to represent indexing by time, $N$ is the number of equal-length time steps captured in the dataset, and the subscript $i$ indexes a particular ABM simulation. Similarly, we define an output of our EBM as a time series $\hat{M}_i \in \hat{M}$:
$$
   \hat{M}_i \coloneq \{ \hat{X}^{\left(0\right)}_i, \dots, \hat{X}^{\left(N\right)}_i \},
$$
with the $\hat{\cdot}$ notation denoting output from the EBM.
Given an error metric (e.g. mean squared error or similar), the optimization problem is therefore:
\begin{equation} \label{eq:closure}
\begin{split}
\underset{\theta}{\text{arg~min}} &~~ \sum_{i=1}^{I} \sum_{k=1}^{N}  ||X^{(k)}_i - \hat{X}^{(k)}_i||_2^2 \\
   ~  \text{s.t.~~} X_i^{(k+1)}\ & = \ODESolve\left(F,X_i^{(k)},t,t + \Delta t,\theta\right), \\
   ~  \hat{X}_i^{\left(0\right)} &= X_i^{\left(0\right)} .
\end{split}
\end{equation}

When paired with a gradient-based optimizer, such as ADAM, the parameters of the model $\theta$ can be tuned to minimize this error. A complete definition of the optimization problem of Eqn. \ref{eq:closure} requires a graph definition, definition of features and states, and specific forms for $\phi$ and $\beta$. These will vary depending on the modeling task. Section \ref{sec:results} details all model specifics for our motivating case study.

\section{Case Study: Population Migration in Response to Flooding in Baltimore} \label{sec:results}

\begin{figure*}[]
        \centering
            \begin{overpic}[width=1.0\linewidth]{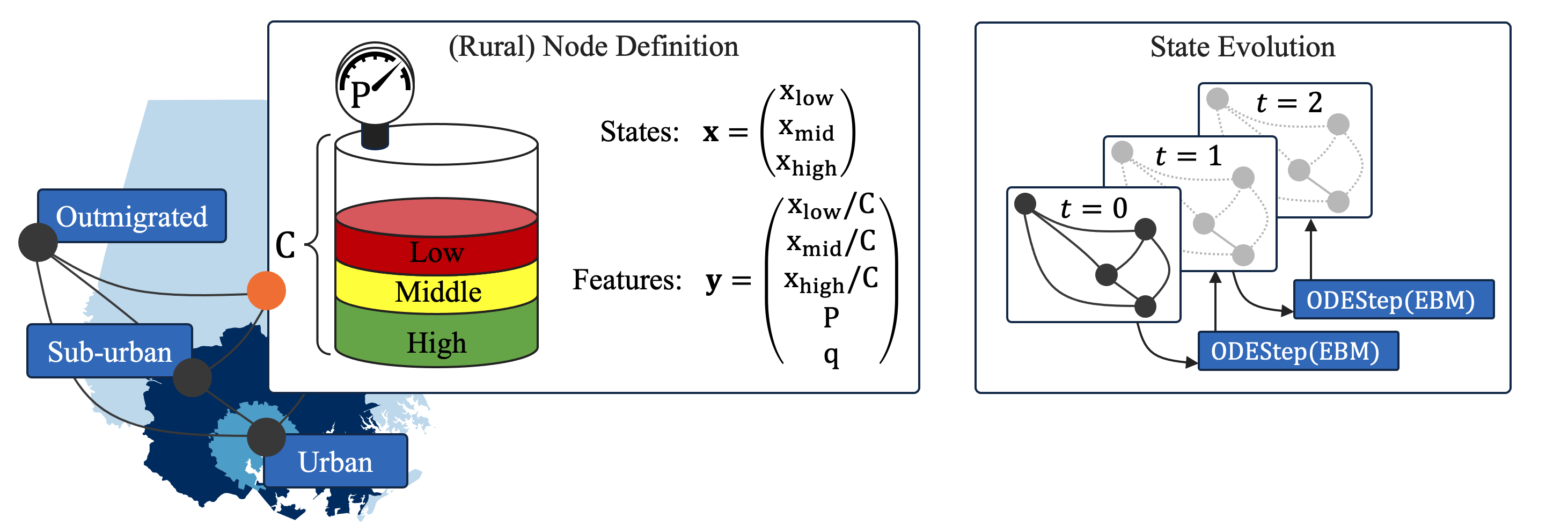}  
            \put(17.5,30){(a)}
            \put(63,30){(b)}
	    \end{overpic}  
	    \caption{After aggregation into nodes in a graph, one can associate states and features to each node. In (a), the definition for the nodes is depicted. Each has capacity for housing, denoted $C$. A mixture of the three sub-populations (low, middle, and high income groups) occupy a fraction of the available housing. The node's states are defined by the composition of the sub-populations. The nodal features are attributes that are specified or derived from model states. Here, they are mixture fraction, a notion of pressure, and a latent variable $q$ to act as a `catch-all' feature. The EBM evolves the states with an ODE solver, as shown in (b). This evolution procedure is differentiable, meaning the parameters of the model can be automatically tuned to minimize an objective function with a gradient-based optimizer.}
		\label{fig:fig_02}
\end{figure*}

\subsection{ABM: Coupled Human-Natural City Evolution - Coastal (CHANCE-C)}

\subsubsection{Model Goals and Description}

In this work, we use the CHANCE-C ABM framework \cite{yoon2023structural} to model space-time socioeconomic dynamics in Baltimore City and County to highlight the capabilities of coarse-grained models to capture dynamic emerging patterns.  

The CHANCE-C spatial domain is comprised of 1220 U.S. census-derived block groups with varying population densities. Household agents can occupy a representative property, each accounting for 100 real-world housing entities and sharing identical structural and neighborhood characteristics within a given block group. This results in roughly 15k household agents. Each household agent is parameterized with a simple utility function that accounts for the perceived utility of each available representative property as a function of structural properties, neighborhood quality, and flood hazard. Each agent is also parameterized with a housing budget, calculated as a function of their household income (based on block-group level U.S. Census data), that determines their purchasing power. At each time step, the developer agent identifies the block groups where demand exceeds the supply and increases the supply by 5\% along with a 5\% price increase for all properties. For the block groups that have not experienced any exceeding demand in 5 years, the developer agent decreases the price by 5\%.

Each time-step in CHANCE-C represents a year, where the housing dynamics comprise the following sequence of events:  
\begin{itemize}
\item New household agents are created (population influx),
\item A percentage of household agents move (property vacancy),
\item Each household agent identifies ten available properties within their budget, calculates the utility of each of them, and ranks them by descending order of utility values,
\item The available properties are matched with interested agents where the agents with the highest utility for a property and the highest income among competing agents get priority,
\item The developer agent assesses the market and adjusts supply and price, and finally
\item Agents that are not assigned a property at the end of the process outmigrate.
\end{itemize}

Flood risk is assessed as an avoidance behavior: a certain percentage of agents are assumed to exclude flood-prone properties from their housing search \cite{yoon2023structural}. 

\subsubsection{Dataset}
A dataset of 50 unique 50-year-duration CHANCE-C model simulations was constructed for this case study. Each simulation contained the same model assumptions; that is, that agents employ a simple avoidance utility for evaluating flood risk in home purchase and relocation decisions. Each of the model runs used a random initial vacancy rate, population growth rate, in-migrating population income percentile, and building growth rate. Each of these model attributes modifies exogenous factors - the core model mechanics do not change across the model runs. On a mid-grade laptop, the mean ABM simulation time is 295 seconds, the majority of which is consumed by modeling housing market dynamics. If run sequentially, this represents a wall-clock time of about 4 hours for dataset generation. The 50-simulation dataset was divided into training (50\%), validation (15\%), and test (35\%) data splits.

\subsubsection{Coarse-Graining} \label{sec:aggregation}
The output of the CHANCE-C ABM is a year-by-year time history of agent block group locations. In a coarse-graining step, we project the agent-based dynamics onto a coarse representation of nodes and sub-populations.

We choose to spatially aggregate the spatial domain of the ABM (Census block groups) into three representative zones: urban, sub-urban, and rural, to be categorized by a distance to the central business district. A fourth zone is added as a placeholder for tracking agents that leave the domain / outmigrate. For each time step of the ABM output, agents are classified into one of these four zones.

We similarly choose to aggregate agents into one of three categories: low income, middle income, or high income. The thresholds for each of these categories are based on successive tertiles of the distribution of agent incomes. For this work, agent income classification is fixed for each ABM model run. 

After coarse-graining, the ABM output is reduced to 12 degrees of freedom -- four nodes times three sub-populations on each node -- reported over 50 years with 1-year temporal resolution. Time series data for in-migrating population, $G_{\nu_i}(t)$, and aggregated nodal housing capacity, $C_{\nu_i}(t)$, are similarly captured and used as exogenous inputs for the EBM.

The classification thresholds for the aggregation are listed in Tab. \ref{tab:classification}.  Note that the choice of aggregation is somewhat arbitrary and can be readily modified, e.g. varying the spatial or socioeconomic granularity, definitions for the aggregated zones, etc.

\begin{table}

\begin{subtable}{1\columnwidth}
\sisetup{table-format=-1.2}   
\centering
   \begin{tabular}{@{} l*{3}{S} @{}}
      \toprule
      & {Urban} & {Sub-urban} & {Rural}  \\ 
      \midrule
      Rule & {$d \leq 0.126$} & {$0.126 < d \leq 0.355 $} & {$d>0.355$}  \\ 

      \bottomrule
   \end{tabular}
   \caption{Spatial aggregation for distance to central business district, $d$. }\label{tab:spatial}
\end{subtable}

\bigskip
\begin{subtable}{1\columnwidth}
\sisetup{table-format=-1.2}   
\centering
   \begin{tabular}{@{} l*{3}{S} @{}}
      \toprule
      & {Low income} & {Middle income} & {High income}  \\ 
      \midrule
      Rule & {$I \leq \$26.5\text{k} $} & {$\$26.5\text{k} < I \leq \$36.7\text{k} $} & {$I > \$36.7\text{k} $}  \\ 
      \bottomrule
   \end{tabular}
   \caption{Agent classification for income $I$. }\label{tab:income}
\end{subtable}

\caption{CHANCE-C simulation output aggregation rules} \label{tab:classification}
\end{table}

\subsection{Problem formulation}
With a model granularity specified by the aggregation procedure of Section \ref{sec:aggregation}, one now needs to define (i) the nodal features, (ii) an appropriate surrogate function for pressure, $\phi$, and (iii) the scaling function $\beta$. 

Note that in many machine learning contexts, feature extraction, learning, selection, and associated tasks play a critical role in a model's performance. Here, to promote model interpretability, we explicitly define each node's feature vector as the concatenation of population mixture fraction, nodal `pressure' , and a catch-all unconstrained latent feature (Fig. \ref{fig:fig_02}a):
\begin{equation}
    \mathbf{y}_{\nu} = \{ \mathbf{x}_{\nu}/C_{\nu} \oplus P_{\nu} \oplus q_\nu \} \in \mathbb{R}^5,
\end{equation}
where the $\oplus$ operator denotes vector concatenation, $C_\nu$ is the housing capacity at node $\nu$, and $q$ is the latent variable to be tuned during optimization. 
The 'pressure' $P_\nu$ is defined as:
\begin{equation}
    P_\nu = \frac{1}{C_{\nu}}\sum \mathbf{x}_\nu .
\end{equation}
The scaling function $\beta$ is:
\begin{equation}
    \beta\left(x_{\nu_i},x_{\nu_j}\right) = \frac{1}{C_{\nu_i}C_{\nu_j}}\mathbf{x}_{\nu_i}\left(1 - \sum \mathbf{x}_{\nu_j}\right) .
\end{equation}
Lastly, we substitute an artificial neural network for $\phi$. The neural network is a small multi-layer perceptron. The architecture is 3 layers; the first two with a width of 5 neurons (corresponding to the dimension of the input) with `swish' activation functions. The last layer is a linear layer of width 3, corresponding to the output dimension of the network.

For simplicity, we use a first-order Euler integration scheme for updating model states:
\begin{equation}
    X^{(k+1)} = X^{(k)} + \Delta t \times F\left(X^{(k)}, Y^{(k)}, t; \theta\right).
\end{equation}

\subsection{Model training and selection}

The model has a total of 82 parameters: 78 for the artificial neural network, and 1 learnable latent feature for each node. These parameters are collected into the parameter vector $\theta$ for the optimization problem of Eqn. \ref{eq:closure}.

\begin{figure}[]
    \centering
\includegraphics[width=1.0\linewidth]{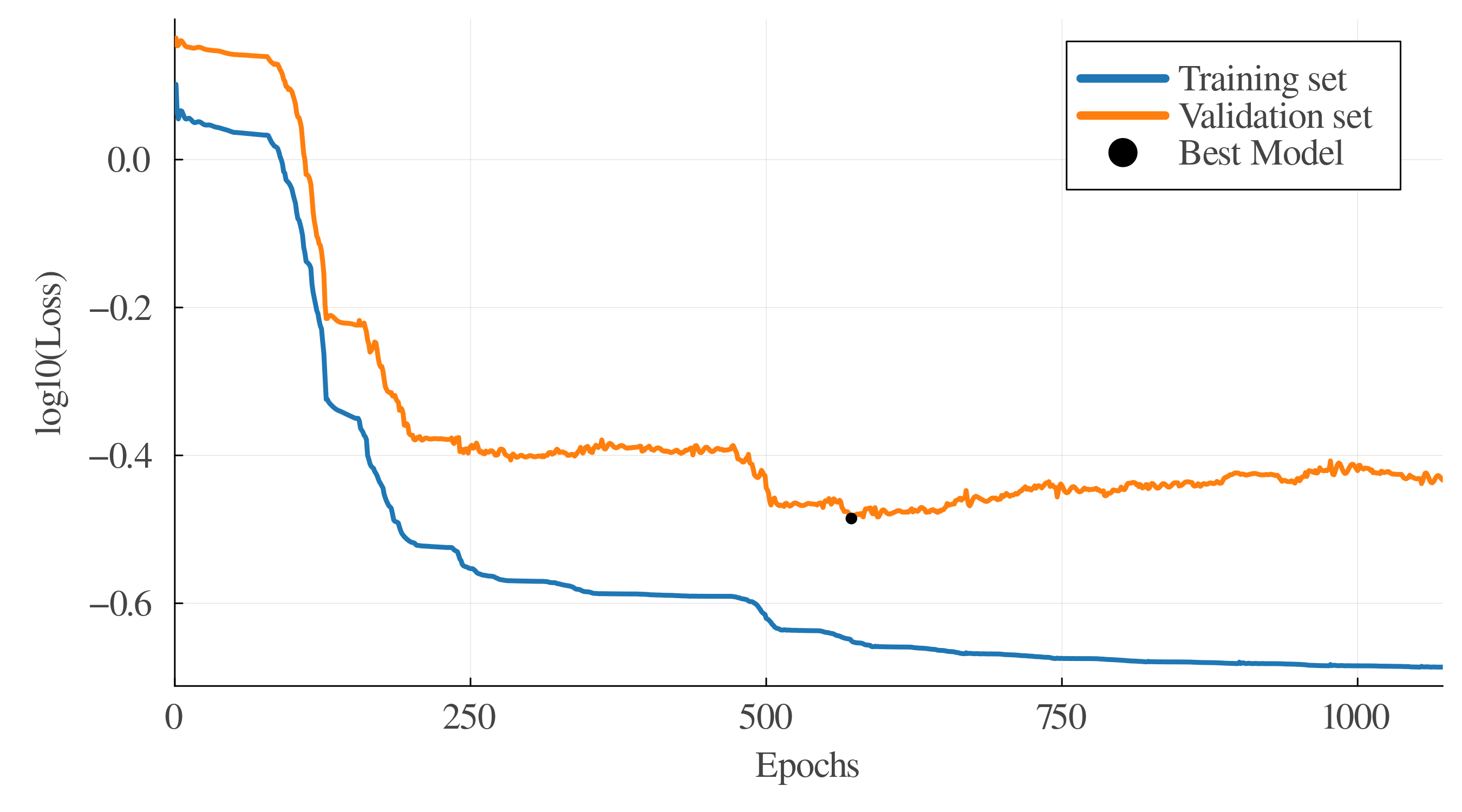}
    \caption{Model training history. `Best Model' is selected as the weights that minimize the loss for the validation set. }
    \label{fig:fig_03}
\end{figure}

Training is performed in a straightforward manner; given a convergence criteria and a training and validation data split, model outputs are evaluated against the datasets according to a loss function (absolute squared error in this case). When converged, the best model weights are selected to be those that minimize the error of the validation set. The convergence criteria for this exercise is when the loss of the validation set reaches a local minimum for at least 500 epochs. For this work, we use the Adam optimizer with a learning rate of 0.01. EBM training according to this model selection criteria concluded in about 1200 epochs. The training history is shown in Fig. \ref{fig:fig_03}.

\subsection{Results}

\begin{table}[] 
\centering
\begin{tabular}{@{} l*{66}{S} @{}} 
  \toprule
  & {MAPE} & {Best} & {Worst} & {MAE} & {Best} & {Worst}  \\ 
  \midrule
  Train & {$11.9\%$} & {$9.61\%$} & {$16.0\%^*$} & {$22.4$} & {$14.6^*$} & {$37.0$}  \\ 
  Val. & {$11.4\%$} & {$9.85\%$} & {$14.8\%$} & {$25.5$} & {$15.8$} & {$38.6^*$}  \\ 
  Test & {$11.9\%$} & {$10.1\%$} & {$14.7\%$} & {$21.8$} & {$15.4$} & {$31.3$} \\ 
  \bottomrule
\end{tabular}
\caption{Summary of performance metrics for the trained model across the training, validation, and testing datasets. The metrics are Mean Absolute Percent Error (MAPE) and Mean Absolute Error (MAE) reported in units of thousands of people. $~^*$These exemplar runs are visualized in Appendix \ref{app:results}.}
\label{tab:results}
\end{table}

\begin{figure*}[]
    \centering
    \begin{overpic}[width=1.0\linewidth]{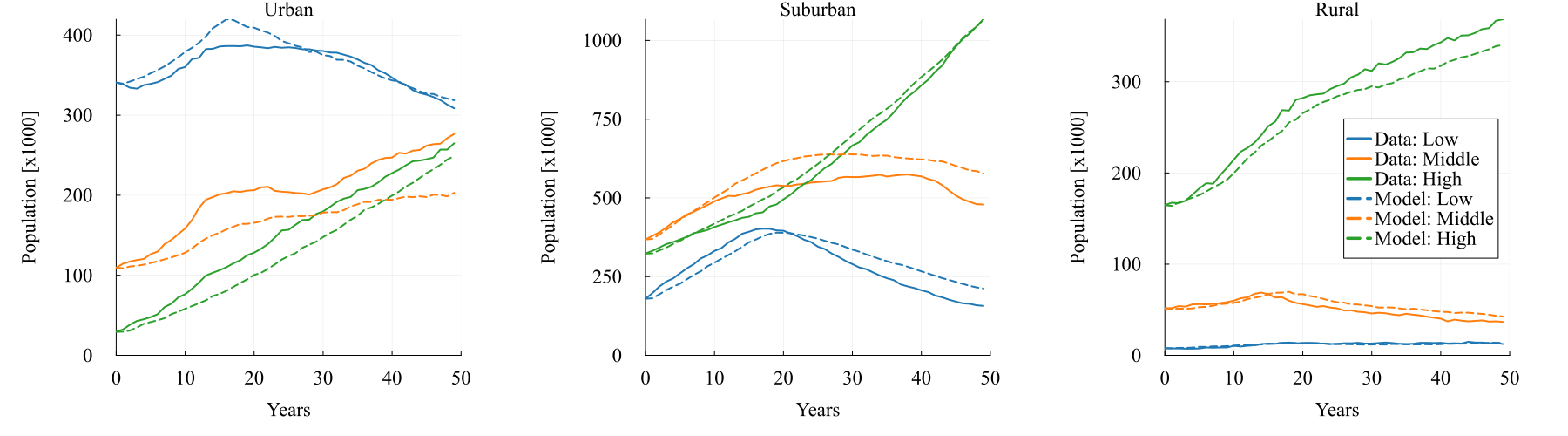} 
    \end{overpic}
    \caption{A protoypical experiment output (taken from test set). Population dynamics on the nodes exhibit complex, nonlinear behaviors. The EBM is able to qualitatively reproduce these trends for each data trace. Over the duration of the experiment, the EBM has a 10.5\% Mean Absolute Percentage Error.}
    \label{fig:fig_04}
\end{figure*}

\begin{figure}[]
    \centering
    \begin{overpic}[width=1.0\linewidth]{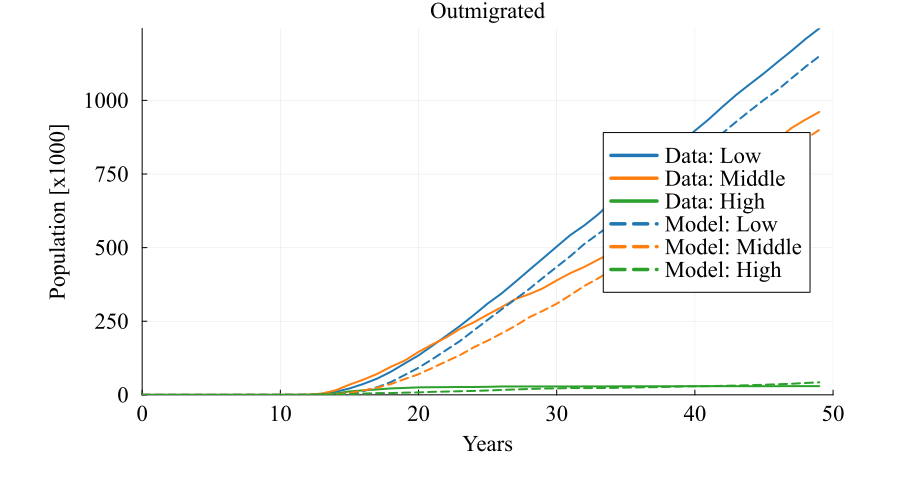} 
    \end{overpic}
    \caption{In addition to tracking flux between Urban/Sub-Urban/Rural nodes, the EBM can also track outmigration. Shown here is the onset of outmigration at approximately 10 years into the model simulation.}
    \label{fig:fig_05}
\end{figure}

A prototypical experiment is shown in Figs. \ref{fig:fig_04} and \ref{fig:fig_05} alongside the ground-truth time series data. In this experiment -- pulled from the testing dataset -- the dynamics of the populations at the nodes are nonlinear, as evidenced by the non-monotonicity of the data trajectories and the magnitudes of their derivatives. Each of the Urban, Sub-urban, and Rural nodes experience population loss to the `Outmigrated' node, though this effect is most pronounced in low-income urban inhabitants, which experience a population loss of about 100k people (about 25\%) from the peak. The high-income group is least vulnerable to outmigration among the sub-populations. These trends persist through all 50 ABM simulations, albeit to varying degrees depending on the simulation parameterization.

The EBM captures these trends, with good qualitative agreement between the true behavior of each sub-population and the model output. Quantitatively, this EBM experiment achieved 10.5\% Mean Absolute Percentage Error (MAPE) and a Mean Absolute Error (MAE) of 25.1k people. The overall performance of the EBM across the entire dataset is summarized in Tab. \ref{tab:results}. Interestingly, the EBM is also able to accurately predict the onset and magnitude of outmigration, as shown in Fig. \ref{fig:fig_05}. These curves represent the cumulative flux of people out of Baltimore County. Until approximately year 10, there is no outmigration. At a point of criticality (e.g. exceeding a model-learned threshold), the network can no longer support the population, triggering the outmigration.

In addition to the experiment of Fig. \ref{fig:fig_04}, we additionally visualize poor-performing models in the Appendix for completeness.

The average wall-clock simulation time for the trained EBM on identical hardware is 6.1ms. This represents a speedup of a factor of 50k over that of the ABM.

\section{Discussion} \label{sec:discussion}
In this work, we formulated, constructed, and demonstrated a novel proof-of-concept graphical Neural ODE for forecasting coarse-grained socioeconomic dynamics. Our procedure involved a data aggregation step, construction of a graph time series (complete with nodal states and features), construction of an EBM, and tuning the EBM to match the coarse-grained data. This EBM is a reduced-order differentiable surrogate model for the ABM output. In this section, we highlight and elaborate on two key concepts of this study: (i) model extensibility, and (ii) model utility.

\subsection{Extensibility and Resolution}
Our EBM has a specific functional form that restricts the model's behavior to exclusively population flux on a graph. The model relates the magnitude of the flux values to the pairwise differences of nodal features. Additionally, the nodal features are manually prescribed. These modeling choices act to dramatically restrict the range of possible behaviors of the EBM. While restrictive, the good qualitative and quantitative performance of the model confirms that these modeling choices were appropriate for our task.

Some situations may require a more expressive modeling abstraction to faithfully capture system dynamics. Our EBM approach is readily extensible to handle varying granularity, including number of nodes in the graph, the dimensionality of the nodal states, and the definition of nodal features. Here, the model resolution was arbitrarily chosen to highlight the capability of the model to extract pertinent relationships at a particular resolution. Depending on modeling goals, one may need to shift up or down the spectrum of model complexity (in terms of degrees of freedom) to be able to resolve certain phenomena.

Furthermore, in this study we did not perform a feature selection task, nor did we search over inductive priors to select a best-performing model. Including more pertinent features at the nodes may help model performance. For this case study, these features could include the distance to coast or other body of water, historical inundation or flood depth data, some characterization of local infrastructure, and the like. An intermediate feature selection -- or generation -- task could similarly improve performance, where a mapping from system observables to features is inserted in the ML pipeline; e.g. $\mathbf{y} = \psi(X,Y;\theta)$. In this context, in addition to learning system dynamics in the form of an EBM, the model is also learning to extract the most relevant features from the data.

\subsection{Utility}
We utility of the presented method is multifold: (i) computational speed-up, (ii) ABM calibration through shared parameterization, (iii) hybrid ABM-EBM, and (iv) coarse-grained `Digital Twins' of real systems.

\subsubsection{Surrogate Modeling}
Through coarse-graining, we have reduced the dimensionality of the system by a factor of hundreds or even thousands, depending on the scale of the particular ABM. Simulating the reduced-dimension system through an ODE call is computationally inexpensive for `reasonable' EBM resolutions (e.g. $n_z \times n_x < 100$). This allows for greater capacity for simulation surveys and performing searches over previously intractable parametric spaces.

\subsubsection{ABM Calibration}
It is possible to construct an ABM-EBM pair that has a shared parameterization; that is, all parameters that define a particular ABM are also explicitly defined in the differentiable EBM. By virtue of this shared parameterization, using the surrogate model, one can perform ABM calibrations and sensitivity studies without needed to perform an ensemble of ABM experiments.

\subsubsection{Hybridized ABM-EBM}
Often the bottleneck in ABM simulations is performing agent-agent interactions, sometimes iteratively. Such can be the case in modeling markets (e.g. housing market, as implemented in CHANCE-C). Here, these low-level ABM interactions are coarse-grained into an equivalent tuned EBM. While there is significant information loss in this coarse-graining step and the associated EBM, one can architect a hybrid ABM-EBM to regain some of the lower level interactions. 

As an example, consider our present case study. Given a trained EBM, one can straightforwardly implement a rule-based developer agent that can choose to add housing capacity on any of the nodes based on that node's states and features. In this manner, the ABM-EBM can still be used to quickly evaluate traditional modeling tasks typically reserved for pure ABMs.

\subsubsection{Digital Twins}
This work has so far assumed that the proposed EBMs are surrogates of ABMs. The same modeling techniques can be applied to real-world data, akin to the `Digital Twin' modeling paradigm. In this setting, the trained EBM can be considered a differentiable surrogate model for an area of interest. The manner in which one can interact with the EBM is not changed; e.g. one can similarly construct a hybrid ABM-EBM to perform `what-if' scenarios catered for an in-real-life area of interest. Combined with model differentiability, this opens up opportunities to explore optimal control and policy crafting in a wide variety of scenarios.

\section{Conclusion} \label{sec:conclusion}
We have presented a novel methodology for constructing and tuning graph-based Neural Ordinary Differential Equations (Neural ODEs) for coarse-grained representations of socioeconomic systems. By integrating machine learning techniques with differential equations, our approach provides a tractable, interpretable, and efficient alternative and compliment to traditional high-fidelity agent-based models (ABMs). 

Our model captures the essential dynamics of socioeconomic phenomena on a graph, preserving critical system behaviors while significantly reducing model complexity. Through a case study focused on migration under flood aversion in Baltimore, we demonstrated the ability of our model to reproduce complex socioeconomic dynamics with good qualitative and quantitative agreement.

The advantages of our approach include: 
\begin{itemize}
    \item Computational efficiency: the reduced-dimensionality of our EBM translates to faster simulations, enabling broader parametric studies and more extensive `what-if' analysis,
    \item Interpretability: the coarse-grained, equation-based nature of our model maintains interpretability, allowing for insights into the underlying mechanisms driving socioeconomic dynamics,
    \item Extensibility: our methodology is flexible and can be adapted for various levels of granularity, additional features, and hybridized modeling approaches, making it versatile for different use cases, and
    \item Differentiability: the differentiable surrogate model opens up possibilities for advanced techniques like optimal control and policy crafting, providing a powerful tool for decision-making in socioeconomic planning and resilience strategies.
\end{itemize}

This approach holds promise for real-world applications where traditional high-resolution models are computationally infeasible. By creating coarse-grained digital twins of actual systems, our methodology can directly inform policy-making, disaster management, and urban planning, enhancing our ability to understand and respond to complex socioeconomic phenomena.

\bibliographystyle{IEEEbib}
\bibliography{main}

\null\newpage
\appendix
\onecolumn
\section{Exemplars}\label{app:results}

\begin{figure*}[ht]
    \centering
    \begin{overpic}[width=0.9\linewidth]{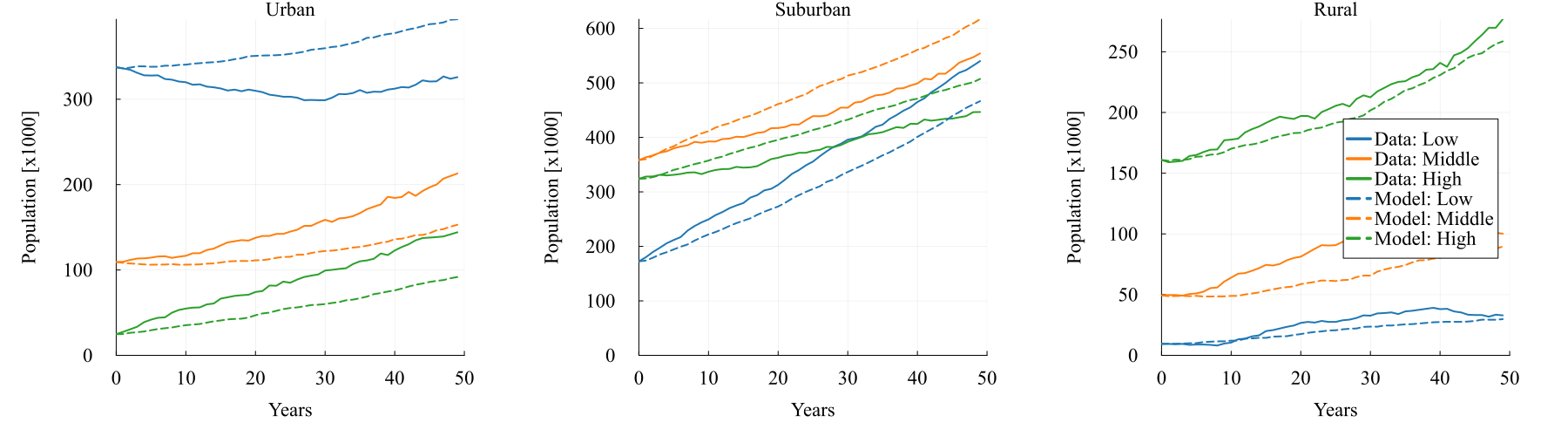} 
    \end{overpic}
    \caption{Run \# 1: MAPE: 16.0\%, MAE: 28.8k. (Worst MAPE)}
    \label{fig:worst_mape}
\end{figure*}

\begin{figure*}[ht]
    \centering
    \begin{overpic}[width=0.9\linewidth]{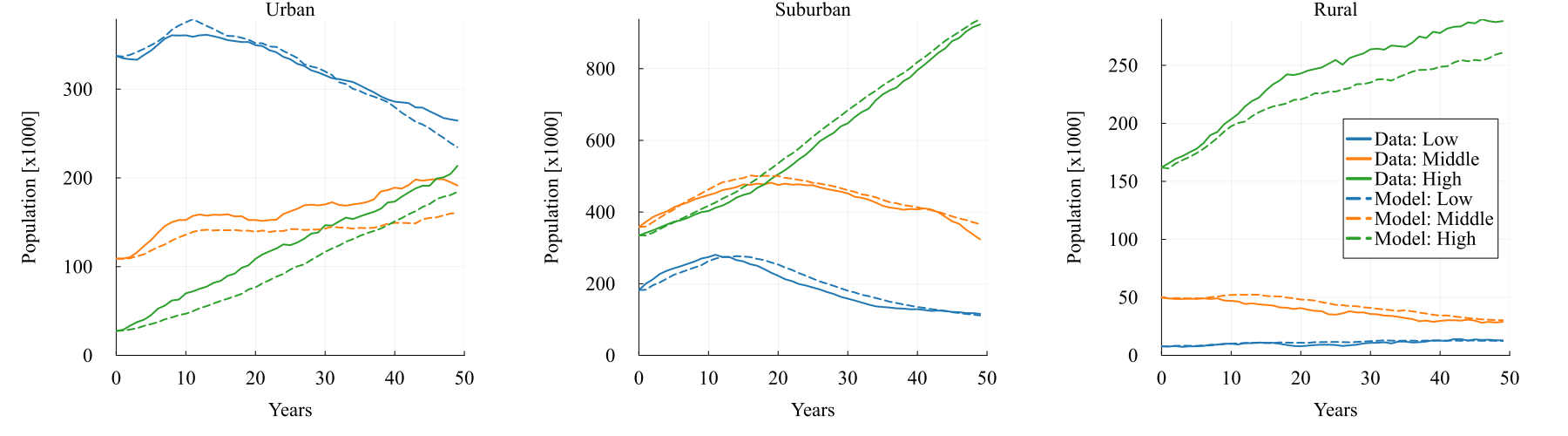} 
    \end{overpic}
    \caption{Run \# 38: MAPE: 9.42\%, MAE: 14.6k. (Best MAE)}
    \label{fig:best_mae}
\end{figure*}

\begin{figure*}[ht]
    \centering
    \begin{overpic}[width=0.9\linewidth]{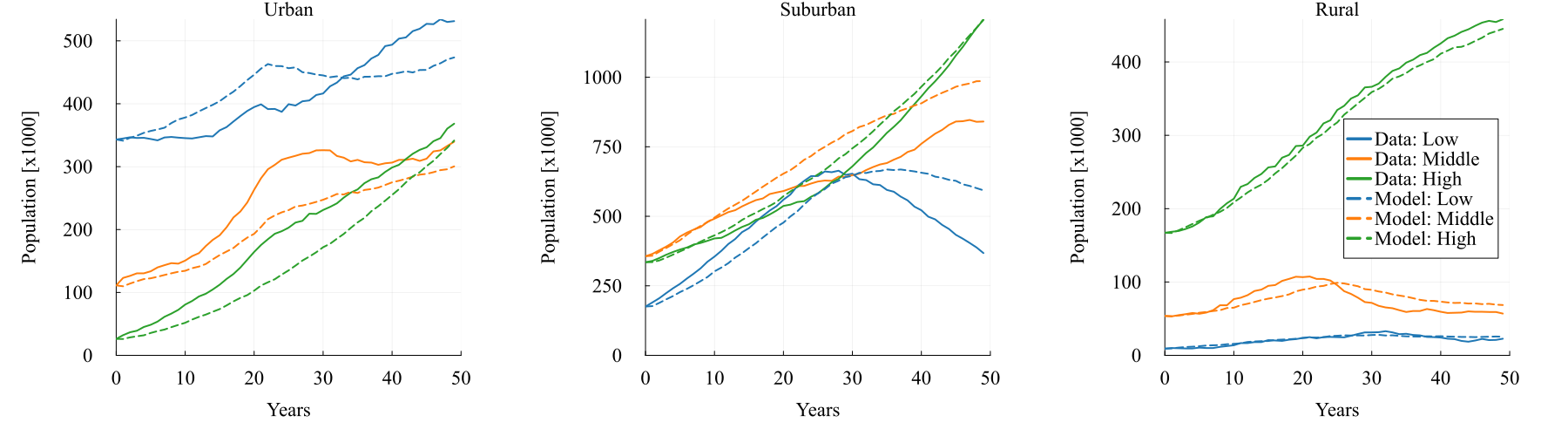} 
    \end{overpic}
    \caption{Run \# 45: MAPE: 13.0\%, MAE: 38.7k. (Worst MAE)}
    \label{fig:worst_mae}
\end{figure*}

\end{document}